\documentclass[journal]{IEEEtran}
\usepackage{amsmath,amsfonts}
\usepackage{algorithmic}
\usepackage{algorithm}
\usepackage{array}
\usepackage[caption=false,font=normalsize,labelfont=sf,textfont=sf]{subfig}
\usepackage{textcomp}
\usepackage{stfloats}
\usepackage{url}
\usepackage{verbatim}
\usepackage{graphicx}
\usepackage{siunitx}
\usepackage{cite}
\usepackage{bm}     % \bm
\usepackage{dsfont} % \mathds
\usepackage{booktabs} % toprule midrule bottomrule
\usepackage{colortbl} % table colors
\usepackage{multirow} % \multirow
\usepackage[nolinks=true]{hyperref} % remove hyper-linkd
\usepackage{silence} % remove warnings from hyperref
\usepackage{doi} % for doi in references
\usepackage{amsthm}
\usepackage{soul} % barred text
\usepackage{tikz}
\AddToHook{shipout/foreground}{%
    \ifnum\value{page}=1
    \begin{tikzpicture}[remember picture,overlay]
        \node[
            anchor=center,
            align=center,
            font=\scriptsize,
            text width=1.2\textwidth
        ] at ([yshift=1cm]current page.south) {%
            This work has been submitted to the IEEE for possible publication. 
            Copyright may be transferred without notice, after which this version 
            may no longer be accessible.
        };
    \end{tikzpicture}
    \fi
}

\definecolor{tabledark}{gray}{0.89}
\definecolor{tablelight}{gray}{0.93}
\definecolor{tablelighter}{gray}{0.97}

\definecolor{commentcolor}{RGB}{201,94,0} % dark orange

\newcommand{\quotes}[1]{``#1''}

\let\oldcite\cite
\renewcommand{\cite}[1]{%
  \ifthenelse{\equal{#1}{}}{\textcolor{red}{[cite]}}{\oldcite{#1}}%
}

\newtheorem{theorem}{Theorem}
\newtheorem{assumption}{Assumption}

\def \papertitle{Vanishing Contributions: A Unified Framework for Smooth and Iterative Model Compression}

\begin{document}

\title{\papertitle}

\author
{
	Lorenzo~Nikiforos,~\IEEEmembership{Graduate Student Member,~IEEE,}
    Luciano~Prono,~\IEEEmembership{Member,~IEEE,}
    Charalampos~Antoniadis,~\IEEEmembership{Member,~IEEE,}
    Fabio~Pareschi,~\IEEEmembership{Senior Member,~IEEE,}
    Riccardo~Rovatti,~\IEEEmembership{Fellow,~IEEE,}
	and Gianluca~Setti,~\IEEEmembership{Fellow,~IEEE}% <-this % stops a space
    \thanks{Code available at: \url{https://github.com/foros15/vanishing-contributions}}%
    \thanks{L. Nikiforos, L. Prono, and G. Setti are with the Department of Electronic and Telecommunication, Politecnico di Torino, 10129 Torino, Italy.
    (e-mail: \{lorenzo.nikiforos, luciano.prono, gianluca.setti\}@polito.it.}
    \thanks{C. Antoinadis is with King Abdullah University of Science and Technology (KAUST), Saudi Arabia.
    (e-mail: charalampos.antoniadis@kaust.edu.sa).}%
    \thanks{F. Pareschi is with the Department of Electronic and Telecommunication, Politecnico di Torino, 10129 Torino, Italy, and also with the Advanced Research Center on Electronic Systems (ARCES), University of Bologna, 40125 Bologna, Italy.
    (e-mail: fabio.pareschi@polito.it)}
    \thanks{R. Rovatti is with the Department of Electrical, Electronic, and Information Engineering, University of Bologna, 40136 Bologna, Italy, and also with the Advanced Research Center on Electronic Systems (ARCES), University of Bologna, 40125 Bologna, Italy.
    (e-mail: riccardo.rovatti@unibo.it).}%
}

\markboth{Nikiforos \MakeLowercase{\textit{et al.}}: \papertitle}{}

% Remember, if you use this, you must call \IEEEpubidadjcol in the second

\maketitle

\begin{abstract}
The increasing scale of Deep Neural Networks (DNNs) introduces the need for compression techniques such as pruning, quantization, and low-rank decomposition. While these methods are very effective at reducing memory, computation, and energy consumption, they may introduce severe accuracy degradation, which is often mitigated by using iterative, gradual compression.
However, different compression techniques require distinct iterative approaches, and some result in unstable, discontinuous model fine-tuning.

We introduce Vanishing Contributions (VCON), a unified framework for the smooth, iterative transition of DNNs into a compressed form.
Rather than replacing the original network directly with its compressed version, VCON executes both in parallel during fine-tuning. The contribution of the original (uncompressed) model is progressively reduced, while that of the compressed model is gradually increased. This affine combination allows the network to slowly adapt, improving stability and mitigating accuracy degradation.

We evaluate VCON on computer vision and natural language processing benchmarks, using multiple compression strategies.
In most settings, our framework improves accuracy over post-shot and iterative baselines. Typical gains exceed 1\%, while some configuration exhibits improvements above 15\%.
VCON is thus compatible with existing compression techniques and consistently improves performance across diverse tasks.

\end{abstract}


\section{Introduction}

\IEEEPARstart{D}{eep Neural Networks} (DNNs) have shown significant capabilities in solving a wide array of complex tasks across multiple domains. These tasks include, but are not limited to: computer vision, natural language processing, speech recognition, medical image analysis, autonomous driving, and even drug discovery~\cite{lecunDeepLearning2015, wangSurveyDeployingMobile2022}. Their remarkable performance has made DNNs the state of the art for many of these tasks. 

However, DNNs come with a major drawback: their models require substantial computational resources, memory, and energy. This high resource consumption makes their deployment in resource-constrained environments challenging, such as on mobile devices, in edge computing, or in real-time systems~\cite{wangSurveyDeployingMobile2022,chengSurveyDeepNeural2024}.

Different studies have shown that DNN models are strongly overparameterized~\cite{allen-zhuLearningGeneralizationOverparameterized2019}, meaning that the number of parameters and the complexity are much larger than necessary to achieve optimal performance. It has been shown that a significant portion of the network's parameters or complexity can be removed without a substantial loss in performance~\cite{hanDeepCompressionCompressing2016}. The reduction approaches, commonly referred to as DNN compression, play a crucial role in making these models more efficient, both in terms of memory usage and computational power. The goal of DNN compression is to create lighter, more efficient models that retain as much of the original model's performance as possible.

Several techniques have been developed to achieve DNN compression. Model pruning, parameter and data quantization, and layer-level compression techniques are among the most important and widely used approaches.
\begin{figure}
  \centering
  \includegraphics[width=\linewidth]{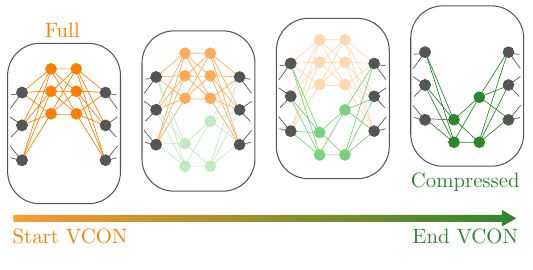}
  \caption{Illustration of VCON (structured pruning): from left to right, the original block (orange) slowly disappears while the contribution of the compressed block (green) is gradually incorporated.}
  \label{fig:vanishing_expl}
\end{figure}

DNN pruning involves selectively removing certain parameters, neurons, filters, or entire layers from the network~\cite{chengSurveyDeepNeural2024, vaderaMethodsPruningDeep2022}. The selection process is typically based on scoring mechanisms that quantify the importance of each network component and help identify which parts of the network are redundant or have a minimal impact on performance. For example, neurons or connections with weights close to zero are often considered unimportant and can be pruned. By pruning unnecessary parts of the network, the model's overall size is reduced, resulting in lower memory usage and computational load.

Another class of techniques for DNN compression relies on quantization, i.e., reducing the precision of the network's parameters. By using fewer bits to represent each parameter, the memory usage of the model is reduced, and its computational cost is decreased~\cite{guoSurveyMethodsTheories2018, wengNeuralNetworkQuantization2023}. This process involves converting standard 32-bit floating-point values to lower-bit-width representations, such as 8-bit floating-point or even 4-bit integers. An extreme case is the use of single-bit binary weights.

Lastly, the use of alternative data representations is common practice, including weight sharing, multiplexing, and Low-Rank Decomposition (LRD)~\cite{mishraSurveyDeepNeural2020}. The goal is to obtain a model that closely replicates the behavior of the original while representing its structure more efficiently.
These methods can significantly reduce the network's memory footprint, enabling it to run more efficiently on devices with limited resources.

Despite their effectiveness in reducing memory footprint and computational cost, many DNN compression methods lead to suboptimal performance. In particular, compressed models often suffer from significant accuracy degradation or fail to generalize as well as their original counterparts, especially when aggressive compression techniques are applied.
To alleviate this, the literature proposes a large plethora of approaches that \emph{iteratively} and \emph{gradually} compress DNN during the training/fine-tuning phase~\cite{gusakAutomatedMultiStageCompression2019, zhouIncrementalNetworkQuantization2017, tanakaPruningNeuralNetworks2020, gordienkoAdaptiveIterativePruning2019, tanDropNetReducingNeural2020, panProgressiveChannelShrinkingNetwork2024, tianHigherRanksAdversarial2023, liaoEmbeddingCompressionIsotropic2020, mengGradientAwareIncrementalNetwork2025, kholiavchenkoIterativeLowRankApproximation2019}. However, each compression approach requires a different strategy, which makes the design of training algorithms very complex. Additionally, some of these approaches introduce significant computational overhead, and the continuous modifications inherent to many gradual compression schemes can destabilize training, underscoring the need for more robust and efficient gradual compression strategies. 

To cope with this, we propose a technique called Vanishing Contribution (VCON), a unified approach that enables a gradual transition from the original network to its compressed version. Instead of directly replacing the original model with the compressed one, we place both in parallel during fine-tuning. The contribution of the original (uncompressed) model is progressively reduced by scaling its output, allowing a smooth, controlled shift toward the compressed model. Fig.~\ref{fig:vanishing_expl} shows a simple diagram illustrating the intuition behind the proposed method.
Although this method was first proposed in~\cite{pronoMultiplyAndMaxMinNeuron2025} for a very specific case, this work expands the original idea by demonstrating a wide range of alternative applications.

We demonstrate the versatility of this approach, which can be applied across various compression contexts and data domains. Specifically, we test the model with pruning, binary quantization, and LRD both for computer vision and natural language processing tasks.
VCON is thus proven to be a generalizable method that yields consistent gains across multiple benchmarks and is easy to implement. During fine-tuning, VCON temporarily runs the original and compressed blocks in parallel, but it does not modify the final network structure produced by conventional compression techniques.

The remainder of this paper is organized as follows. First, we provide an overview of iterative methods in Section~\ref{sec:related_works}, as they are based on concepts closely related to VCON. Then, Section~\ref{sec:compression_methods} presents a detailed analysis of various compression techniques proposed in the literature, with particular emphasis on those adopted in this work. In Section~\ref{sec:vanishing}, we thoroughly explain the formulation of VCON. Then, we demonstrate in Section~\ref{sec:experimental} the effectiveness of VCON against a representative set of progressive compression methods, across a wide range of applications. Section~\ref{sec:limit_future} outlines the limitations of the proposed technique and directions for future research. Finally, we draw the conclusions.

\section{Related Works}
\label{sec:related_works}

Several works in the literature explore DNN compression, proposing strategies to reduce model size while maintaining acceptable performance. Among them, a prominent class of techniques leverages iterative compression, progressively compressing the model during or across multiple training phases. 

Iterative compression is based on the principle that a network can better adapt to structural changes when they are introduced gradually rather than all at once. Instead of applying a single, strong compression operation, iterative techniques apply small, incremental modifications, allowing the model to retrain and recover at each stage. This incremental process enables the network to redistribute important information across the remaining parameters, effectively maintaining its representational power despite progressive simplifications.

The literature on iterative compression provides examples that work with most compression techniques, such as pruning, quantization, and LRD. Iterative pruning gradually removes the least important weights, inserting retraining phases between two subsequent pruning steps to recover the lost performance~\cite{tanakaPruningNeuralNetworks2020, gordienkoAdaptiveIterativePruning2019, tanDropNetReducingNeural2020}. For example, in~\cite{panProgressiveChannelShrinkingNetwork2024}, the authors propose an approach that iteratively updates channel importance estimates via an affine combination during training, gradually identifying and pruning redundant channels in a stable manner. In another work~\cite{tianHigherRanksAdversarial2023}, authors introduce Rank-based PruninG (RPG), an iterative weight pruning method guided by a rank-based objective that encourages higher matrix ranks for weights. It involves alternating phases in which interconnections are removed and in which are reintroduced, updating masks based on weight importance and gradients, and fine-tuning the sparse model.

Similarly, iterative quantization progressively reduces the precision of selected subsets of weights, starting from high-precision formats and moving toward lower-bit representations, while retraining the model after each quantization stage~\cite{liaoEmbeddingCompressionIsotropic2020}. For example, in~\cite{mengGradientAwareIncrementalNetwork2025}, the authors propose an iterative quantization approach that gradually quantizes neural network weights by classifying them based on gradient-based importance. Low-importance weights are quantized first, followed by fine-tuning, and the process is repeated to achieve significant compression while preserving accuracy. Alternatively, some work quantizes only a few parameters in a sparse way, first quantizing the most important ones -- using pruning-like scoring metrics -- then gradually decreasing quantization sparsity until all the parameters are quantized~\cite{zhouIncrementalNetworkQuantization2017}.

In the case of LRD, weight matrices are incrementally approximated with lower-rank factors, and retraining steps allow the network to adapt to the reduced parameterization~\cite{kholiavchenkoIterativeLowRankApproximation2019}. In all these cases, the progressive nature of the compression, coupled with intermediate fine-tuning, helps preserve model accuracy despite substantial reductions in size and complexity.

Similar to the aforementioned approaches, the VCON method offers an alternative approach based on the same principle of gradually transitioning from the original model to the compressed one. However, they operate along fundamentally different dimensions.
Iterative compression reduces the number of parameters in discrete stages. At each step, a subset of parameters is compressed, and the model is fine-tuned to recover performance before the next compression stage. This process progressively shrinks the model by explicitly removing parameters.
In contrast, VCON does not immediately discard parts of the original model. Instead, it gradually reduces their contribution by decreasing their magnitude over time, while simultaneously increasing the influence of the compressed counterpart, improving training stability. %Rather than eliminating parameters outright, VCON smoothly shifts the computational load from the original model to the compressed one.

Moreover, VCON is agnostic to the employed compression method, resulting in a flexible, unified framework and the advantage of using a single, simple compression framework. By standardizing the iterative compression process, VCON can facilitate the development of novel approaches, particularly those where gradual compression would otherwise be architecturally complex or computationally expensive.

\section{Compression Methods Overview}
\label{sec:compression_methods}

In this section, we provide an overview of existing compression methods compatible with the VCON technique, our general compression framework that smoothly transitions a network from full-size to compressed blocks during fine-tuning.
We stress that VCON can, in principle, be applied to other compatible approaches as well.

\subsection{Pruning}

Pruning is generally performed following two main strategies: structured~\cite{heFilterPruningGeometric2019, yuNISPPruningNetworks2018, luoThiNetFilterLevel2017, liPruningFiltersEfficient2017}, and unstructured pruning~\cite{sunSimpleEffectivePruning2024, liuGroupFisherPruning2021, tangManifoldRegularizedDynamic2021, frantarSparseGPTMassiveLanguage2023}. The former involves removing neurons, filters, or entire blocks. This approach is hardware-friendly, meaning it does not require hardware adaptation to maintain computational performance. However, it typically removes a smaller portion of weights to preserve comparable accuracy to that of the original network. 
On the contrary, unstructured pruning focuses on removing individual interconnections, regardless of their position in the network. This results in a higher percentage of weights being removed than in the structured case, leading to a significant reduction in model size but with the drawback of a more complex hardware implementation.

The selection of the parts to be removed is typically guided by a scoring system. Each removable entity (e.g., interconnections, neurons, filters, or blocks) is assigned a score reflecting its importance within the network. Entities are then removed in ascending order (from the least to the most important) until the desired pruning rate is achieved.

Pruning methods based on scoring can be applied at different levels of granularity, yielding three pruning groups: global, layer-wise, and N:M. In global pruning, scores are computed for all entities across the entire network, and the lowest-scoring entities are removed, regardless of the block/layer they belong to. In layer-wise pruning, scores are ranked and pruned within each layer separately, ensuring a more balanced distribution of weight sparsity. A third type, namely N:M pruning, introduces a structured constraint: within every local group of M interconnections, exactly N are kept, and the remaining are pruned.

To simplify the experimental analysis, we restrict our study to strategies that rely on magnitude-based scoring. These methods assume that interconnections with the smallest absolute weights -- or the smallest norm of the weights associated with a neuron -- contribute less to the network's overall output and can therefore be removed.

\subsection{Low-Rank Decomposition}
\label{sec:overview_lrd}

LRD reduces both memory consumption and computational cost, and can be applied to both convolutional and fully connected layers~\cite{denilPredictingParametersDeep2014, lebedevSpeedingupConvolutionalNeural2015, liConstrainedOptimizationBased2018, sainathLowrankMatrixFactorization2013, zhangAcceleratingVeryDeep2016, novikovTensorizingNeuralNetworks2015}. In the case of fully connected layers, the weight matrix can be approximated as the product of two lower-rank matrices.
More in detail, let $\bm W \in \mathds{R}^{n \times m}$ be a weight matrix. LRD is represented by two matrices $\bm A \in \mathds{R}^{n \times r}$ and $\bm B \in \mathds{R}^{r \times m}$ such that $r \ll \min(n,m)$. With this decomposition, we aim to replace $\bm W$ with the product $\bm A\bm B$.
The computational cost and the memory complexity of a fully connected layer are $O(mn)$, whereas its LRD has complexity $O(r(m+n))$. This approach offers a benefit when $r(m+n)<mn$, i.e., if $r < \frac{m+n}{mn}$.

In this work, we employ the truncated Singular Value Decomposition (SVD), a common approach for obtaining low-rank matrices. 
Specifically, matrix $\bm W$ can be factorized as $\bm W = \bm U\bm\Sigma \bm V^T$, where 
$\bm U \in \mathds{R}^{n \times n}$ and $\bm V \in \mathds{R}^{m \times m}$ are unitary matrices, and $\bm\Sigma \in \mathds{R}^{n \times m}$ is a rectangular diagonal matrix whose diagonal elements are the singular values arranged in descending order. The truncated SVD keeps only the $r$ largest singular values. The factorized matrices $\bm U$, $\bm\Sigma$ and $\bm V^T$ become $\bm{\hat U} \in \mathds{R}^{n \times r}$, $\bm{\hat\Sigma} \in \mathds{R}^{r \times r}$ and $\bm{\hat V}^T \in \mathds{R}^{r \times n}$.
From this, we finally set $\bm A=\bm{\hat U}$ and $\bm B=\bm{\hat\Sigma} \bm{\hat V}^T$.
After factorization, fine-tuning is typically required to achieve accuracy comparable to that of the original model.

\subsection{Quantization}

Quantization is a widely used technique for compressing DNNs, reducing the precision of weights and activations.
In its general form, quantization maps each full-precision weight or activation $x \in \mathcal{Q}$ to a value $\tilde{x} \in \tilde{\mathcal Q}$, where $\mathcal Q$ and $\tilde{\mathcal Q}$ are finite sets and $|\tilde{\mathcal Q}| < |\mathcal Q|$, i.e., the number of representable values for $\tilde{x}$ is smaller than for $x$.
The size of each set is determined by the bit-width -- i.e., the number of bits used to encode the information.
For instance, 8-bit quantization provides 256 distinct levels, while lower bit-widths yield smaller sets.

Quantization techniques can be broadly classified into post-training quantization, where quantization is applied after training, and quantization-aware training, where the model is trained while accounting for quantization effects, typically resulting in higher final accuracy.  

Finally, binary and ternary quantization~\cite{tuAdaBinImprovingBinary2022, vargasBiPerBinaryNeural2024, chenTernaryLLMTernarizedLarge2024, xuTerViTEfficientTernary2022} are extreme forms of quantization that constrain weights and/or activations to only two or three distinct values, respectively, drastically reducing model size and simplifying arithmetic operations.
With binary quantization, weights are constrained to $\{-1, +1\}$. Given a real-valued weight $w \in \mathds{R}$, the binary quantized value $w_b$ can be obtained through the simple sign function:
\[
w_b = \text{sign}(w) =
\begin{cases}
+1, & \text{if } w \geq 0 \\
-1, & \text{otherwise}.
\end{cases}
\]

Although binary networks achieve significant reductions in memory and computational cost, they result in a non-negligible degradation in accuracy. 
To mitigate this problem, each group of binary weights or activations is multiplied by a fixed scaling factor to restore their original overall scale and compensate for limited precision~\cite{rastegariXNORNetImageNetClassification2016}.
Scaling factors can be learned during training or predefined, depending on the method used. Our experiments include binary quantization with a fixed scaling factor.

\subsection{Training schedule and constraints}

Training schedules play a key role in model compression and are directly relevant to the experimental evaluation of VCON. In our study, we consider methods that either fine-tune the model after compression (post-shot), iteratively increase model compression, or incorporate compression constraints during training via the Straight-Through Estimator (STE), as these approaches are either used as baselines or within VCON itself.
Post-shot fine-tuning~\cite{mozaffariSLiMOneshotQuantization2025, huOPQCompressingDeep2022} involves a fine-tuning phase that follows the compression of the model. This allows the model to recover some of the accuracy lost during compression.

On the other hand, iterative compression~\cite{changIterativeClusteringPruning2023, yeProgressiveDNNCompression2019, panProgressiveChannelShrinkingNetwork2024} progressively reduces model complexity over multiple training iterations. At each step, the model is compressed slightly and then fine-tuned. This gradual process allows the model to adjust more effectively to compression, reducing the performance drop compared to one-shot and post-shot methods. VCON generalizes this process and achieves similar adaptive compression within a single training schedule. In fact, it gradually shifts the contribution from full-size blocks to compressed blocks, thereby emulating the effect of iterative compression.
STE~\cite{bengioEstimatingPropagatingGradients2013} is commonly used to incorporate non-differentiable compression constraints during training, allowing the model to improve the compressed configuration.
As an example, STE is applied to perform the so-called quantization-aware training: quantization is applied dynamically during the forward pass, while during the backward pass, parameters are restored to their full-precision version~\cite{huhStraighteningOutStraightThrough2023}. The same applies to pruning, where, before the forward pass, parts of the model are dynamically removed, but then restored right before the backward pass~\cite{huSSTEContinuousPruning2024}.

\begin{figure}
  \centering
  \includegraphics[width=\linewidth]{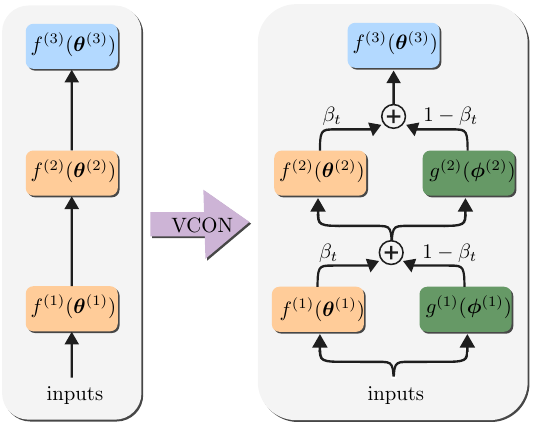}
  \caption{Illustration of block-wise VCON: in this example, the first two blocks $f_{\bm{\Theta}}^{(1)}$ and $f_{\bm{\Theta}}^{(2)}$ are progressively replaced with their compressed counterparts $g_{\tilde{\bm \Theta}}^{(i)}$, while the final block $f_{\bm{\Theta}}^{(3)}$ remains uncompressed.}
  \label{fig:vanishing_multi}
\end{figure}

\section{Vanishing Contribution}
\label{sec:vanishing}

In this section, we provide a detailed explanation of VCON, which enables a smooth transition from an original model to its compressed version by progressively reducing the influence of the former during training while increasing that of the latter.

\subsection{Vanishing contribution definition}

To define VCON, we first introduce a typical setting in model compression. A DNN model is composed of $L$ blocks (e.g., linear or convolutional layers) defined as
\begin{equation}
    \bm x^{(i+1)}=f^{(i)}\left(\bm x^{(i)}; \bm\theta^{(i)}\right)\quad
    \bm x^{(0)} = \bm x,
\end{equation}
where $i \in \{0, \dots, L-1\}$, $f^{(i)}$ is the function describing the $i$-th block behavior, $\bm\theta^{(i)}\in\mathds R^{d_i}$ is the concatenation of all the $d_i\in\mathds N$ parameters of the $i$-th block and $\bm x$ is the input of the model.

We compress a subset of these blocks through transformation operations $G^{(i)}$ as
\begin{equation}
    \left(g^{(i)}, \bm\phi^{(i)}\right) = G^{(i)}\left(f^{(i)}; \bm\theta^{(i)}, \mathcal N\right),
\end{equation}
where $\bm\phi^{(i)}\in\mathds R^{c_i\leq d_i}$ is the concatenation of all the parameters after compression and $g^{(i)}(\bm x^{(i)}; \bm\phi^{(i)})$ describes the compressed block function. We explicitly model the dependence on the entire neural model $\mathcal {N}$, as some compression methods rely on information from the full model. Here, $G^{(i)}$ can be interpreted as pruning, quantization, or LRD.

For notational simplicity, we omit explicit dependence on $\bm x^{(i)}$ and $i$ whenever it can be inferred from the context
\begin{equation}
\begin{array}{l}
    f(\bm\theta) := f^{(i)}(\bm x^{(i)}; \bm\theta^{(i)})\\
    g(\bm\phi) := g^{(i)}(\bm x^{(i)}; \bm\phi^{(i)}).
\end{array}
\end{equation}

Within the VCON framework, the original function $f(\bm\theta)$ is replaced by a new function $h_{\beta_t}(\bm\theta;\bm\phi)$, defined as the affine combination between the original function and its compressed counterpart as
\begin{equation}
    h_{\beta_t}(\bm\theta;\bm\phi) = \beta_t f(\bm\theta) + (1-\beta_t)g(\bm\phi),
\end{equation}
where $\beta_t$ is a numerical sequence in the range $\left[0,1 \right]$, that progressively varies with the training step $t\in \{0,\dots, T-1\}$, where $T\in \mathds N$ is the total number of training steps. In more detail, $\beta_t$ controls the gradual transition from the original function $f(\bm\theta)$ to its compressed version $g(\bm\phi)$ as the training process advances. 

The value of $\beta_t$ is scheduled during the training process through a monotonically non-increasing function, so that $\beta_0 = 1$, yielding $h_1(\bm\theta;\bm\phi)\equiv f(\bm\theta)$, and $\beta_{T-1}=0$, yielding $h_0(\bm\theta;\bm\phi)\equiv g(\bm\phi)$. Fig.~\ref{fig:vanishing_multi} illustrates a visual example.

In this work, we use a piece-wise linear scheduler, defined as
\begin{equation}
    \beta_t = \max\left(1-\frac{t}{Q},0\right),
\end{equation}
which linearly decreases for $Q$ steps, and remains zero for $t \geq Q$ (in this case, $h_0(\bm\theta;\bm\phi)\equiv g(\bm\phi)$ for $t \geq Q$).
This process facilitates a smooth and simultaneous transition by allowing independent compression of each block. 

\subsection{VCON as an optimum tracker}

Let us define the implicit objective function for each VCON block as 
$\mathcal L_{\beta_t}(\bm\theta; \bm\phi)$, whose value directly depends on $h_{\beta_t}(\bm\theta; \bm\phi)$. This implies that the objective function evolves smoothly as $\beta_t$ varies, so that the family $\mathcal L_{\beta_t}(\bm\theta, \bm\phi)$ can be interpreted as a homotopy between objectives corresponding to different values of $\beta_t$. For notational convenience, we fix $\bm\theta$ and we define $\mathcal L_{\beta_t}(\bm\phi) := \mathcal L_{\beta_t}(\bm\theta; \bm\phi)$.

Then, let $\bm\phi \mapsto \bm\phi_t$ be the concatenation of parameters at each training step $t$, with training update
\begin{equation}
    \bm\phi_{t+1} = \bm\phi_t - \eta\nabla\mathcal L_{\beta_t}(\bm\phi_t),
\end{equation}
where $\eta$ is the learning rate. The following implications can be easily extended if $\bm\theta$ is updated, too. 

Finally, let us define the minimizer $\bm\phi_{\beta_t}^* = \arg\min_\phi\mathcal L_{\beta_t}(\phi)$ and the distance to the target as $E_t=\mathcal L_{\beta_t}(\bm\phi_t)-\mathcal L_{\beta_t}(\bm\phi_{\beta_t}^*)$.

\begin{assumption}[L-smoothness in $\bm\phi$]\label{ass:smooth}
For all $\beta \in [0,1]$, the function $\mathcal L_\beta(\bm\phi)$ is $L$-smooth in $\bm\phi$, i.e.,
\begin{equation}
\|\nabla \mathcal L_\beta(\bm\phi) - \nabla \mathcal L_\beta(\bm\phi')\|
\leq L \|\bm\phi - \bm\phi'\| \quad \forall \bm\phi,\bm\phi'.
\end{equation}
\end{assumption}

\begin{assumption}[Lipschitz continuity in $\beta$]\label{ass:beta}
For all $\bm\phi$ and all $\beta,\beta' \in [0,1]$, the objective varies smoothly with $\beta$, i.e.,
\begin{equation}
\|\mathcal L_\beta(\bm\phi)-\mathcal L_{\beta'}(\bm\phi)\|
\leq K \|\beta - \beta'\|.
\end{equation}
\end{assumption}

\begin{theorem}\label{thm:descent}
Under Assumptions~\ref{ass:smooth} and~\ref{ass:beta}, the distance to the target satisfies the following approximate descent inequality:
\begin{equation}
E_{t+1} \leq E_t - \frac{\eta}{2}\|\nabla\mathcal L_{\beta_t} (\bm\phi_t)\|^2 + 2K\|\beta_{t+1}-\beta_t\|.
\end{equation}
\end{theorem}

The full proof is provided in Appendix~\ref{app:proof}. This result suggests that $E_{t+1}$ closely tracks $E_{t}$ if the variation of $\beta_t$ is sufficiently small from one training step to the next. 
When we do not apply a gradual compression, we get in one shot from $f(\bm\theta)$ to $g(\bm\phi)$, so $2K\|\beta_{t+1}-\beta_t\|=2K$. In this case, compression introduces an arbitrarily large error, making it challenging to recover the original performance.
When we apply a gradual compression with VCON, $2K\|\beta_{t+1}-\beta_t\|\ll2K$, the potentially large, uncontrolled error is transformed into a sequence of controlled perturbations.

\subsection{An intuitive perspective on VCON working principle}

\begin{figure}
    \centering
    \includegraphics[width=3.5in]{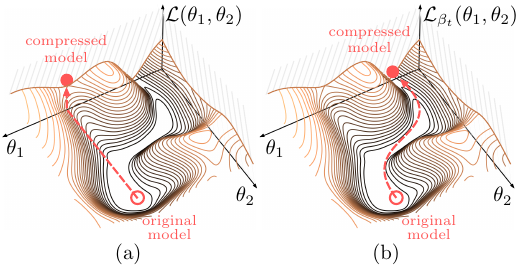}
    \caption{Visual intuition of VCON: when a model parameter is removed abruptly (a), the working point is projected directly onto the hyperplane defined by the remaining dimensions, which is suboptimal. In contrast, if the parameter is removed gradually (b), the working point shifts slowly toward the hyperplane of the remaining dimensions and the model is continuously updated, allowing a greater chance to reach a better local minimum.}
    \label{fig:intuition}
\end{figure}

To further support the rationale for VCON, we propose an intuitive view of model compression. Fig.~\ref{fig:intuition} visually illustrates the proposed intuition. Let us have a model with two parameters, $\theta_1$ and $\theta_2$, of which one is to be zeroed (removed) via pruning. Then let us define the objective function as $\mathcal L(\theta_1, \theta_2) : \mathds R^2 \rightarrow \mathds R$. Removing parameter $\theta_2$ is equivalent to restricting the domain of $\mathcal L$ to $\{(\theta_1, 0)\in \mathds R^2\}$, \emph{abruptly} projecting the current working point to axis $\theta_1$. This adjustment can be far from optimal, even if we linearize the model's behavior and choose to prune the parameter with the smallest local influence on the cost function. Conversely, using VCON, the working point is \emph{gradually} shifted towards the $\theta_1$ axis, yielding $\mathcal L_{\beta_t}(\theta_1,\theta_2)=\mathcal L(\theta_1,\beta_t\theta_2)$. The model is then updated according to the full profile of $\mathcal L$. Our intuition is that this gradual reduction leaves room for the model to adapt its trajectory in the cost function landscape. This means that the model has a better chance of finding a local minimum than it would under an abrupt structural change, even when fine-tuning is performed after compression. Fig.~\ref{fig:intuition} explicates this as a potential scenario where one-shot compression brings the model to a poor local minimum, whether VCON keeps it within the basin of attraction of the original minimum.
This vision can be easily extended to the generic multi-dimensional case, where removing a parameter results in projecting the working point onto the hyperplane defined by the remaining axes.

\subsection{The effect of varying $Q$}
We perform a first experiment to analyze the influence of $Q$ variation on the training dynamics and model performance.
To do so, we evaluate unstructured pruning on the ViT-T/16~\cite{dosovitskiyImageWorth16x162020} model using the CIFAR-10 dataset~\cite{krizhevskyLearningMultipleLayers2009}. We apply a layer-wise pruning STE strategy with a compression ratio of 95\% and compare with and without the VCON technique. Details about the dataset and training setup are in Appendix~\ref{app:datasets} and Appendix~\ref{app:hyperparam}, respectively.

\begin{figure}
  \centering
  \includegraphics[width=\linewidth]{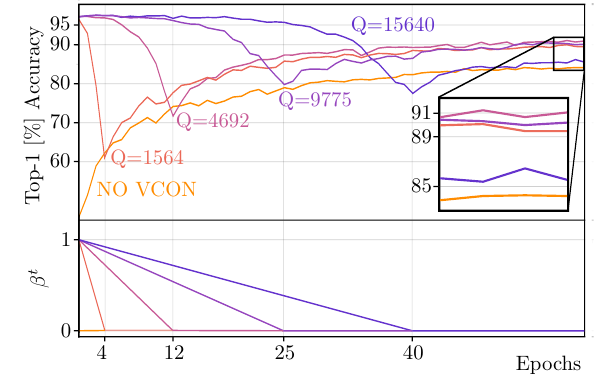}
  \caption{Impact of Training Dynamics with VCON: evolution during training for different transition durations $Q$.}
  \label{fig:training_graph}
\end{figure}

In the VCON setting, while the original blocks gradually fade out, the parallel counterpart progressively takes over and is trained throughout the process. Pruning via STE is naturally applied to the latter, which evolves dynamically during training. This reflects the core idea of VCON: enabling a smooth and flexible transition that integrates seamlessly with existing compression techniques.
We performed the experiment with different values of $Q$, completing the vanishing phase after 4, 12, 25, and 40 epochs ($ Q=1564, 4692, 9775, 15640$), respectively. Fig.~\ref{fig:training_graph} shows the validation-set accuracy curves evolving during the training process.

All the configurations employing the VCON technique consistently outperform the baseline model without VCON, demonstrating the effectiveness of our approach in enhancing model performance. The improvement is evident across all values of the transition hyperparameter $Q$, confirming the general benefit of introducing a smooth transition.

A closer inspection of the training dynamics reveals a recurring pattern. For each value of $Q$, there is a noticeable drop in validation accuracy that aligns closely with the end of the transition period, where the original model blocks are fully faded out.
This temporary drop is likely due to the change in model representation as the compressed path fully takes over. However, in all cases, the model resumes a steady ascent in accuracy, indicating that the network successfully adapts to the new structure once the transition is complete. This aligns with the intuition that the vanishing phase keeps the model near the high-performance region containing the original optimum, which the compressed model subsequently explores to recover performance.

Among the tested configurations, a transition of 12 epochs appears to strike the best balance. This result is in line with Theorem~\ref{thm:descent}. A transition that is too short results in a large drift term $2K\|\beta_{t+1}-\beta_t\|$, resulting in an uncontrolled transition to the compressed model. On the contrary, an excessively long transition results in the model converging prematurely, yielding $\|\nabla\mathcal L_{\beta_t}(\bm\phi_t)\|^2 \approx 0$ and thus $E_{t+1}\approx E_t$.
%A transition that is too short may not allow sufficient time for the compressed model to effectively assimilate the knowledge encoded in the original network, while an excessively long one could delay the learning of the compressed blocks. 

Our findings suggest that intermediate values of $Q$ lead to smoother transitions and better overall performance, highlighting the importance of tuning this hyperparameter to achieve optimal results.

\begin{figure*}
    \centering
    \includegraphics[width=\linewidth]{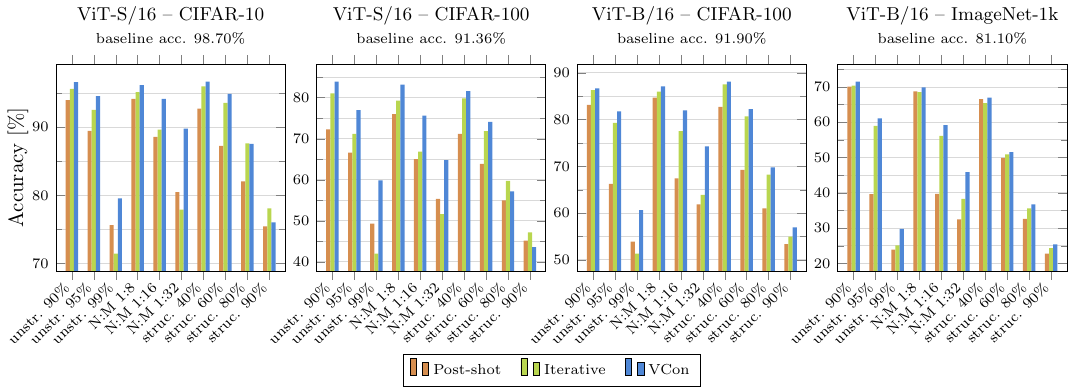}
    \caption{Accuracy on computer vision models compressed and fine-tuned with different STE-based pruning strategies. Plots compare post-shot target compression, gradually increasing pruning ratio (Iterative), and vanishing contributions strategy (VCon).}
    \label{fig:barplot_pruning_cv}
\end{figure*}

\begin{figure*}
    \centering
    \includegraphics[width=\linewidth]{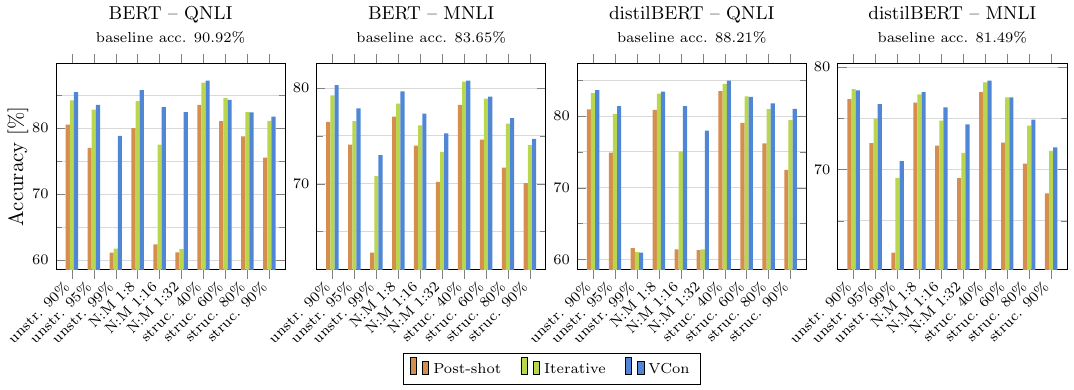}
    \caption{Accuracy on natural language processing models compressed and fine-tuned with different STE-based pruning strategies. Plots compare post-shot target compression, gradually increasing pruning ratio (Iterative), and vanishing contributions strategy (VCon).}
    \label{fig:barplot_pruning_txt}
\end{figure*}

\section{Experimental results}
\label{sec:experimental}

To assess the effectiveness of VCON, we focus on three model compression techniques: pruning, quantization, and LRD. These methods are applied across a range of datasets and models in both computer vision and natural language processing. Specifically, we test a Vision Transformer (ViT)~\cite{dosovitskiyImageWorth16x162020} in its \quotes{Tiny}, \quotes{Small} and \quotes{Base} variants (ViT-T/16, ViT-S/16 and ViT-B/16) on CIFAR-10, CIFAR-100~\cite{krizhevskyLearningMultipleLayers2009} and ImageNet-1k~\cite{dengImageNetLargescaleHierarchical2009} computer vision datasets and two language models, BERT~\cite{devlinBERTPretrainingDeep2019} and distilBERT~\cite{sanhDistilBERTDistilledVersion2020} on Question Natural Language Inference (QNLI)~\cite{demszkyTransformingQuestionAnswering2018} dataset and Multi-Genre Natural Language Inference (MNLI) dataset~\cite{williamsBroadCoverageChallengeCorpus2018}. Further information on datasets is provided in Appendix~\ref{app:datasets}.

\subsection{Baseline Implementation Details}

We evaluate three compression strategies: post-shot-, iterative-, and VCON-based approaches.
In post-shot settings, we apply compression to the target level in a single shot, then fine-tune.
When using VCON, we update $\beta_t$ at each training step.
Similarly, iterative compression progressively updates the compression level at each training step to minimize abrupt changes in the network structure and ensure a transition schedule comparable to VCON.

Different iterative compression approaches are employed for each compression category, each of them updating a compression hyperparameter at each training step. With pruned models, we linearly increase the pruning ratio until it reaches the target value. Iterative compression with LRD is based on a modified version of~\cite{gusakAutomatedMultiStageCompression2019}. Rank is progressively reduced as
\begin{equation}
    r_t = \big\lceil \rho_t (r_\text{max}-r_\text{target})\big\rceil+r_\text{target},\ \rho \in [1,0],\ r_t \in \mathds{N},
\end{equation}
where $\rho$ linearly changes from 1 to 0 for each training step $t$. Each time $r_t \neq r_{t-1}$, we restore the full-rank weight matrix as $\bm{\tilde W}=\bm A \bm B$ (see Section~\ref{sec:overview_lrd}) and then immediately decompose it back to the new rank.
Iterative quantization is performed by applying sparse quantization to a linearly increasing number of weights~\cite{zhouIncrementalNetworkQuantization2017} until all of them are quantized.

All approaches use compression-aware constraints to maximize final accuracy. Specifically, we fine-tune both quantization and pruning with the STE strategy, while with LRD, we directly fine-tune the compressed matrices $\bm A$ and $\bm B$.
Further details on the training are provided in Appendix~\ref{app:hyperparam}.

\subsection{Numerical results}
We test VCON with magnitude-based pruning using three granularity strategies: unstructured layer-wise, unstructured N:M, and structured layer-wise. We evaluate multiple pruning ratios, uniformly removing parameters across layers (self-attention and multi-layer perceptron layers).  Fig.~\ref{fig:barplot_pruning_cv} and Fig.~\ref{fig:barplot_pruning_txt} show pruning results for vision and language models, respectively.

\begin{figure*}[!ht]
    \centering
    \includegraphics[width=\linewidth]{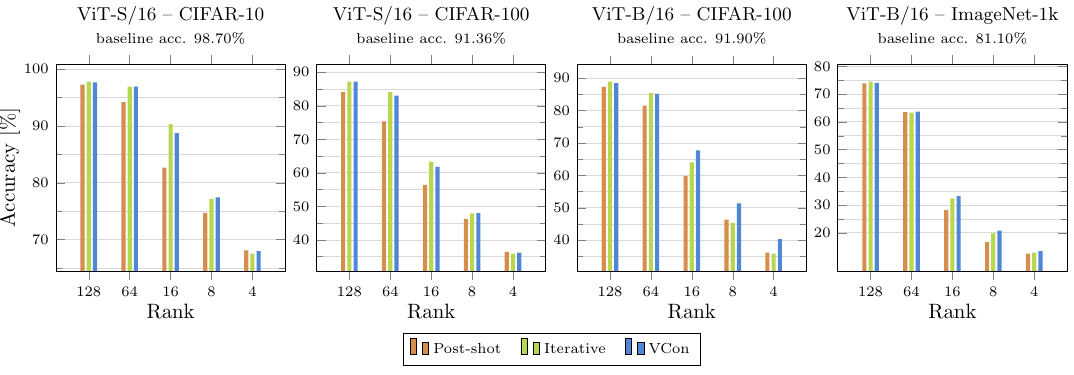}
    \caption{Accuracy on computer vision models compressed with LRD and fine-tuned. Plots compare post-shot target compression, the gradually decreasing rank (Iterative) strategy, and the vanishing contributions (VCon) strategy.}
    \label{fig:barplot_lrd_cv}
\end{figure*}

\begin{figure*}[!ht]
    \centering
    \includegraphics[width=\linewidth]{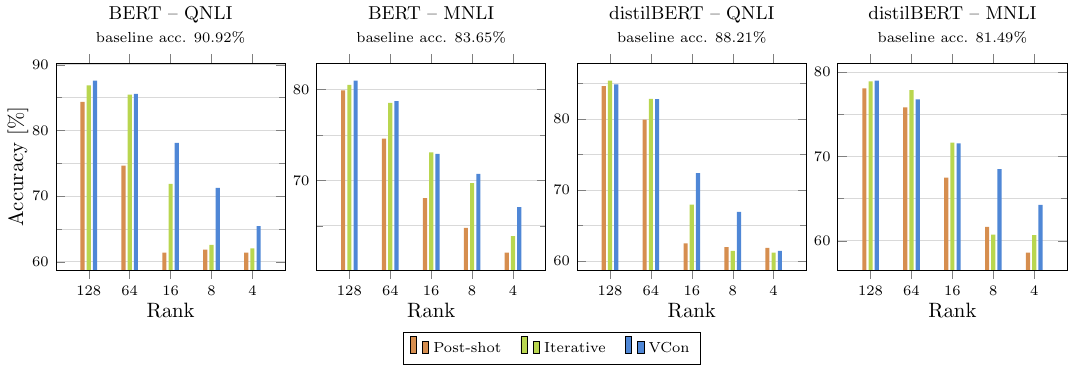}
    \caption{Accuracy on natural language processing models compressed with LRD and fine-tuned. Plots compare post-shot target compression, the gradually decreasing rank (Iterative) strategy, and the vanishing contributions (VCon) strategy.}
    \label{fig:barplot_lrd_txt}
\end{figure*}

Similarly, Fig.~\ref{fig:barplot_lrd_cv} and Fig.~\ref{fig:barplot_lrd_txt} show results with LRD on vision and language models with different rank values.

\begin{figure}
    \centering
    \includegraphics[width=\linewidth]{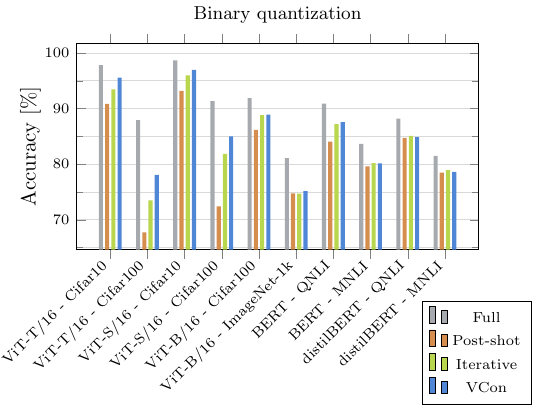}
    \caption{Accuracy on computer vision and natural language processing models, binarized and fine-tuned. Plots compare full-precision model accuracy against post-shot target compression, gradual compression with decreasingly sparse binarization (Iterative), and vanishing contributions strategy (VCon).}
    \label{fig:barplot_quant}
\end{figure}

Finally, we evaluate VCON in the context of binary quantization. In this setting, network weights are constrained to two discrete values, -1 and +1. Rather than introducing learnable scaling parameters, which would require architectural modifications, we use the $\ell_2$ norm of each weight matrix as a scaling factor.
Fig.~\ref{fig:barplot_quant} covers binarization results of both computer vision and natural language processing tasks.

In most settings, our VCON yields consistent improvements in classification accuracy when compared to both the post-shot and iterative pruning baselines. Specifically, most setups show improvements of at least 1\%, while some achieve more than 15\%. Only in some scenarios, VCON results in suboptimal or comparable accuracy.

We observe that VCON's benefits become more pronounced as the compression level grows, but can become less stable under very aggressive compression. In fact, improvements are generally more consistent on smaller models or simpler tasks, suggesting that the smooth transition mechanism is most effective when sufficient model capacity remains relative to task complexity.

Finally, VCON performs slightly worse than the iterative baseline in some LRD setups. However, we note that iterative LRD is computationally expensive, as it requires repeatedly decomposing and recomposing the weight matrices via SVD, which disrupts the model's architecture and computational graph. In contrast, after its initial setup, our framework does not require any architectural modifications during training.

These results further confirm that VCON integrates seamlessly with orthogonal compression strategies within a unified framework.

\section{Limitations and future works}
\label{sec:limit_future}
While the proposed VCON framework demonstrates promising results across multiple tasks and compression techniques, several limitations remain and warrant further investigation.

We observed that certain combinations of models, datasets, and compression methods yield suboptimal or inconclusive results. This suggests that, in some scenarios, baseline compression techniques may already be operating close to their optimal points, leaving limited room for improvement. 
Alternatively, the introduced compression could still undermine training stability -- i.e., even if there is an optimal model configuration, we may not be able to reach it with the current training configuration -- thereby negating the benefits of VCON.
A deeper analysis is required to understand these dynamics and to isolate the conditions under which our method is most effective.
Exploring this direction is an important topic for future work.

Additionally, while we extensively tested VCON on multiple architectures, compression strategies, and task domains, further studies on additional model families and more complex workloads would help broaden its empirical coverage. As an example, future work should extend the evaluation to large-scale generative language models.

Finally, the use of VCON incurs additional computational complexity and memory requirements, since all compressed blocks are duplicated -- even though the overall complexity is lower than some iterative setups, e.g., LRD. This is partially alleviated by freezing full-size blocks $f(\bm\theta)$ -- which, on average, does not affect final accuracy -- to reduce the computational load of back-propagation. Nevertheless, future research should address the increased computational complexity to allow the use of VCON on very large models and resource-constrained training systems.

\section{Conclusion}
\label{sec:conclusion}
In this work, we explored and extensively tested Vanishing Contributions (VCON), a general training strategy that enables a smooth transition from an original DNN to its compressed version. VCON progressively reduces the influence of the uncompressed model during training while the compressed model is gradually activated in parallel.

The proposed VCON technique was evaluated through an extensive set of experiments across three major compression methods: pruning, quantization, and LRD. Experiments were conducted on both computer vision (CIFAR-10, CIFAR-100, ImageNet-1k) and NLP tasks (QNLI, MNLI) using ViT, BERT, and DistilBERT architectures. For pruning, VCON was tested across different sparsity levels using an STE-based approach, with unstructured layer-wise, unstructured N:M, and structured layer-wise granularities. For quantization, the method was tested with an STE-based binary setting. Lastly, LRD experiments were conducted by applying truncated SVD with varying rank constraints. 

Overall, the experimental evaluation confirms that VCON provides a generally superior or at least comparable alternative to other iterative compression strategies.
Moreover, VCON's main strengths lie in its broad compatibility and ease of integration with existing methods. It acts as a unified extension to common training procedures and can be seamlessly combined with state-of-the-art compression pipelines. As demonstrated, this approach improves the robustness and quality of compressed models without requiring complex modifications. 

\appendices
\section{Theorem full proof}
\label{app:proof}
We provide here the full proof of Theorem~\ref{thm:descent}. We start by expanding $E_{t+1}$ by adding and subtracting equal contributions:
\begin{equation}
\begin{aligned}
    &E_{t+1}
    = \mathcal L _{\beta_{t+1}}(\bm\phi_{t+1})
     - \mathcal L_{\beta_{t+1}}(\bm\phi^*_{t+1})\\
    &= \underbrace{\mathcal L _{\beta_{t+1}}(\bm\phi_{t+1})
     - \mathcal L _{\beta_t}(\bm\phi_{t+1})}_{A}
     + \mathcal L _{\beta_t}(\bm\phi_{t+1})
     - \mathcal L_{\beta_{t+1}}(\bm\phi^*_{t+1})\\
    &= A + \underbrace{\mathcal L _{\beta_t}(\bm\phi_{t+1})
     - \mathcal L_{\beta_t}(\bm\phi^*_t)}_{B}
     + \underbrace{\mathcal L_{\beta_t}(\bm\phi^*_t)
     - \mathcal L_{\beta_{t+1}}(\bm\phi^*_{t+1})}_{C}\\
    &E_{t+1} \leq |A| + |B| + |C|.
\label{eq:toterror}
\end{aligned}
\end{equation}

We then bound each term to obtain a bound on $E_{t+1}$.

\subsection{Varying $\beta_t$ with the same parameter set}
By Assumption~\ref{ass:beta}, we get
\begin{equation}
|\mathcal L _{\beta_{t+1}}(\bm\phi_{t+1}) - \mathcal L _{\beta_t}(\bm\phi_{t+1})| \leq K\|\beta_{t+1}-\beta_t\|.
\end{equation}

\subsection{Constant $\beta_t$, distance to objective}
By Assumption~\ref{ass:smooth}, we can apply the gradient descent lemma~\cite{nesterovSmoothConvexOptimization2004}:
\begin{equation}
\mathcal L_{\beta_t}\left(\bm\phi_t-\eta\nabla\mathcal L_{\beta_t}(\bm\phi_t)\right)
\leq \mathcal L_{\beta_t}(\bm\phi_t) - \frac{\eta}{2}\|\nabla\mathcal L_{\beta_t}(\phi_t)\|^2.
\end{equation}
Substituting $\bm\phi_{t+1} = \bm\phi_t-\eta\nabla\mathcal L_{\beta_t}(\bm\phi_t)$ and subtracting left and right terms by $\mathcal L_{\beta_t}(\bm\phi^*_t)$ we get
\begin{equation}
|\mathcal L_{\beta_t}(\bm\phi_{t+1}) - \mathcal L_{\beta_t}(\bm\phi^*_t)|
\leq E_t - \frac{\eta}{2}\|\nabla\mathcal L_{\beta_t}(\phi_t)\|^2,
\end{equation}
since $L_{\beta_t}(\bm\phi_{t+1}) \geq \mathcal L_{\beta_t}(\bm\phi^*_t)$ and $E_t = \mathcal L_{\beta_t}(\bm\phi_t) - \mathcal L_{\beta_t}(\bm\phi^*_t)$.

\subsection{Varying objective function}
In this case, we can bound the difference of minima of the objective function as
\begin{equation}
|\min_{\bm\phi}\mathcal L_{\beta_t}(\bm\phi) - \min_{\bm\phi}\mathcal  L_{\beta_{t+1}}(\bm\phi)| \leq \sup_{\bm\phi}|\mathcal L_{\beta_t}(\bm\phi)-\mathcal L_{\beta_{t+1}}(\bm\phi)|.
\end{equation}

By the definition of the minimizers $\bm\phi^*_t$ and $\bm\phi^*_{t+1}$ and by applying Assumption~\ref{ass:beta} to the right term, we get
\begin{equation}
|L_{\beta_t}(\bm\phi^*_t) - L_{\beta_{t+1}}(\bm\phi^*_{t+1})| \leq K\|\beta_{t+1}-\beta_t\|.
\end{equation}

By substituting the terms in \eqref{eq:toterror}, we get
\begin{equation}
E_{t+1} \leq E_t - \frac{\eta}{2}\|\nabla\mathcal L_{\beta_t} \bm\phi_t)\|^2 + 2K\|\beta_{t+1}-\beta_t\|.
\end{equation}
\hfill $\square$

\section{Datasets and preprocessing}
\label{app:datasets}
To evaluate our models across both computer vision and natural language processing tasks, we employed a selection of well-established benchmark datasets. Below, we provide a description of each of them and the preprocessing technique used in our experiments.

In the case of the computer vision tasks, we employed the following datasets:
\begin{itemize}
    \item CIFAR-10~\cite{krizhevskyLearningMultipleLayers2009} is a dataset consisting of \num{60000} color images, each with a resolution of 32×32 pixels and three color channels (RGB). The images are uniformly distributed across 10 distinct classes. The dataset is split into \num{50000} training images and \num{10000} test images.
    \item CIFAR-100~\cite{krizhevskyLearningMultipleLayers2009} is a dataset also containing \num{60000} $32\times 32$ RGB images and 100 classes. For CIFAR-10, the dataset is split into \num{50000} training and \num{10000} test images.
    \item ImageNet-1K~\cite{dengImageNetLargescaleHierarchical2009} is a large-scale dataset composed of approximately 1.28 million training images and \num{50000} validation images, categorized into \num{1000} object classes. Since the ground-truth labels for the official ImageNet test-set are not publicly available, we split the standard validation set evenly into our validation and test sets.
\end{itemize}
We applied a set of preprocessing techniques to all the computer vision datasets considered in this work. We first resize the input images to $224\times 224$ pixels, then normalize them using dataset-specific standard statistics. We then apply data augmentation using RandAugment, provided by the PyTorch framework, and MixUp.

In the case of the language tasks, we employed the following datasets from the General Language Understanding Evaluation (GLUE) benchmarks:
\begin{itemize}
    \item QNLI~\cite{demszkyTransformingQuestionAnswering2018} is a binary classification task. Each example consists of a question and a context sentence extracted from a Wikipedia article. The task is to determine whether the sentence contains the correct answer to the question. The dataset includes \num{104743} training examples, \num{5463} for validation, and \num{5463} for testing. 
    \item MNLI~\cite{williamsBroadCoverageChallengeCorpus2018} contains pairs of sentences composed of a premise and a hypothesis. The task is to determine the semantic relationship between them by assigning one of three labels: entailment, contradiction, or neutral. The training set comprises \num{392702} examples, with validation and test sets each containing \num{9815} examples, further split into matched (in-domain) and mismatched (cross-domain) subsets.
\end{itemize}

\section{Model training hyperparameters}
\label{app:hyperparam}

Vision models are fine-tuned from models pretrained on ImageNet-21k, while language models are first fine-tuned for \num{8} epochs on the target datasets (QNLI, MNLI).
Table~\ref{tab:training_configs} shows the hyperparameters used for the different models and configurations examined in this work. Note that the value of $Q$ for each setup can be inferred from \quotes{VCON or iterative epochs} and the number of training steps per epoch, which depends on the dataset. The same value is applied to non-VCON iterative approaches, defining the number of epochs the method uses to reach the target compression level. Post-shot approaches keep the same training configuration, only ignoring \quotes{VCON or iterative epochs}.

\begin{table}
    \centering
    \caption{Training configurations and hyperparameters}
    \rowcolors{3}{white}{tablelighter}
    \begin{tabular}{l|ll}
        \toprule
        Model             & \begin{tabular}{@{}l@{}}ViT-T/16\\ ViT-S/16\\ ViT-B/16\end{tabular}
                          & \begin{tabular}{@{}l@{}}BERT\\ distilBERT\end{tabular}\\
        \midrule
        Batch size        & 128
                          & 16
                          \\
        Optimizer         & AdamW
                          & AdamW \\
        LR start          & 1e-4
                          & 2e-5 \\
        LR scheduler      & OneCycleLR
                          & \begin{tabular}{@{}l@{}}EarlyStopping\\(patience=5)\end{tabular} \\
        Warmup (epochs)   & \begin{tabular}{@{}l@{}}
                                1 {\tiny(quantization, pruning)} \\
                                6 {\tiny(LRD)}
                            \end{tabular}
                          & -- \\
        LR start (warmup) & \begin{tabular}{@{}l@{}}
                                1e-6 {\tiny(quantization, pruning)} \\
                                1e-4 {\tiny(LRD)}
                            \end{tabular}
                          & --
                          \\
        \begin{tabular}{@{}l@{}}
             VCON or\\
             iterative epochs
        \end{tabular}
               & \begin{tabular}{@{}l@{}}
                                12 {\tiny(CIFAR-10, CIFAR-100)}\\
                                1 {\tiny(ImageNet-1k)}
                            \end{tabular}
                          & 2 \\
        Total epochs      & \begin{tabular}{@{}l@{}}
                                60 {\tiny(CIFAR-10, CIFAR-100)}\\
                                5 {\tiny(ImageNet-1k)}
                            \end{tabular}
                          & 8 \\
        \bottomrule
    \end{tabular}
    \label{tab:training_configs}
\end{table}

\bibliographystyle{bib/myIEEEtran.bst}
\bibliography{bib/bibliography}

\end{document}